# 以矛盾为中心的群体智能涌现模型

焦文品[1]
北京大学计算机学院
高可信软件技术教育部重点实验室（北京大学）
jwp@pku.edu.cn

**摘要**

群体智能的涌现现象在自然界和人类社会中广泛存在，人们一直在探索群体智能涌现现象的根本原因并试图建立通用的群体智能涌现理论和模型，但现有的理论或模型并没有抓住群体智能的本质，因而也就缺乏一般性，难以同时刻画各种各样的群体智能的涌现现象。本文提出了一种以矛盾为中心的通用的群体智能涌现模型，在该模型中，个体的内在矛盾决定了它们的行为和特征，个体因为竞争环境资源而联系在一起并相互交互，个体之间的交互和群势会影响个体内部矛盾及其在群体中的分布，群体智能表现为个体矛盾的特定分布。该模型完整地阐述了群体智能涌现的条件、动力、途径、形式及过程。为了验证该模型的有效性，本文分析和实现了多种群体智能系统。实验结果表明，该模型具有很好的通用性，能够用来刻画各种群体智能的涌现。

# A Contradiction-Centered Model for the Emergence of Swarm Intelligence

Wenpin Jiao
School of Computer Science, Peking University
Key Laboratory of High Confidence Software Technologies (Peking University), MOE
Email: jwp@pku.edu.cn

## Abstract

The phenomenon of emergence of swarm intelligence exists widely in nature and human society. People have been exploring the root cause of emergence of swarm intelligence and trying to establish general theories and models for emergence of swarm intelligence. However, the existing theories or models do not grasp the essence of swarm intelligence, so they lack generality and are difficult to explain various phenomena of emergence of swarm intelligence. In this paper, a contradiction-centered model for the emergence of swarm intelligence is proposed, in which the

[1] 通信地址：北京市海淀区北京大学计算机学院软件研究所，邮编：100871，电话：15311707727，邮箱：jwp@pku.edu.cn。

本文工作受到国家重点研发计划（2023YFC3502900）和基金重大项目（62192731）的资助。

internal contradictions of individuals determine their behavior and properties, individuals are related and interact within the swarm because of competing and occupying environmental resources, interactions and swarm potential affect the internal contradictions of individuals and their distribution in the swarm, and the swarm intelligence is manifested as the specific distribution of individual contradictions. This model completely explains the conditions, dynamics, pathways, formations and processes of the emergence of swarm intelligence. In order to verify the validity of this model, several swarm intelligence systems are implemented and analyzed in this paper. The experimental results show that the model has good generality and can be used to describe the emergence of various swarm intelligence.

## 1. 引言

在自然界和社会中都能观察到一种普遍存在的现象，即一些无智能或简单智能的个体通过聚集与协同而呈现（或涌现）出某种高智能特性，例如蚂蚁觅食、鸟群迁徙、鱼群避险、自由市场等，人们把这种现象称为群体智能[2]。群体智能常常和涌现、集体等概念联系在一起[8]，涌现被认为是（多智能体）系统呈现出高智能特性的过程[19]，具有高智能特性的群体往往呈现出空间、时间或功能上的有序结构[12]。

自从群体智能这一概念提出以来，引起了众多研究人员的关注，已成为计算机、人工智能、经济、社会、生物等交叉学科的研究热点。人们借鉴那些常见的群体智能涌现现象提出了很多群体智能方法（或算法），例如蚁群算法[6]、粒子群算法[9][18]等，用来解决现实世界中的复杂问题。这些方法只要求参与协同的个体具有简单的计算能力和行为能力，就能在没有集中控制和全局模型的情况下，高效、灵活、健壮地求解或优化复杂问题。

但是，这些方法都不具有足够的一般性（或通用性），它们都有各自特定的适用范围，一般仅针对特定的群体才可能有效。人们在运用特定的群体智能方法时，必须先检查群体是否具有该方法所要求的某些特征，群体中的个体是否遵循方法所规定的方式在行动，以及个体的行为是否可追踪等，否则，群体就可能不能涌现出所期望的智能，或者说无法基于该方法来求解或优化群体遇到的问题。例如，要运用蚁群算法，群体中的个体必须能够像蚂蚁那样产生（或者分泌）类似于信息激素的东西，其它蚂蚁能够感知到这些信息激素，并根据信息激素的强弱来强化自己的行为。因此，人们在实现群体智能时，不得不重复试探各种各样的群体智能方法，以便找到最合适的方法。

然后，即使选用了“以为”最恰当的群体智能方法，也不一定能保证群体能涌现出所期望的智能，这是因为现有的群体智能方法只是模拟了某些涌现现象，而没有抓住涌现的本质。人们至今都无法解释群体智能到底是如何涌现的，个体间的交互或协同与群体智能之间到底存在何种联系？即个体的交互为什么能涌现出这样或那样的群体智能特性、以及个体间的交互是否必然会涌现出预期的群体智能？

现有的方法大多具有三个明显的特征：一是认为群体演进的主要动力来源于个体与环境(包括个体之间)之间的相互影响和相互作用；二是群体存在某种共同的实用主义目标；三是个体的行为由一些简单规则驱动[19]。这些方法虽然强调个体之间的协作、以及协作的优化，但它们实际上并不清楚协作的真正源动力是什么、以及协作的方向是什么。也就是说它们不清楚群体之间为什么协作，也无法预测协作到底会产生什么样的（优化）效果。

本质上讲，环境存在于事物（这里指个体）的外围，环境以及个体与环境的相互作用只是影响个体行为的外部条件（哲学上称为外因）。例如，一般都认为大雁迁徙是因为气候发生变化的原因引起的，但实际上，决定它们是否迁徙的根本原因在于它们的生存和繁衍能否

得到保障，包括是否有足够的食物、生命是否受到威胁、能否找到配偶并成功繁衍后代等，而气候的变化只是引起它们迁徙的某种外在因素而已，而非决定性因素。其次，群体涌现的智能只是观察的结果，而不是群体本身就具有的内生目标。例如大雁在迁徙过程中形成特定的队形并不是因为它们预先就想好要形成该队形，而是因为受到外界环境（例如风、温度等）的影响，它们为了更安全、更省力地飞抵目的地自然而然形成的。第三，外部环境的复杂性和不确定性使得影响个体行为的外在因素数不胜数，并且个体也不可能对环境有完整的感知和认知，人们根本无法完备地把握个体和环境之间的关系，进而准确地定义个体与环境之间的相互影响和相互作用，并获取或定义出完备的行为规则。已有的方法错误地将环境变化以及个体与环境之间的交互作为群体智能涌现的主要动力，这自然无法正确地刻画和建模群体智能，建立的模型也难以适用于现实世界中复杂环境下的群体。

哲学（主要是辩证法）上认为，外因（即外部环境因素）只是影响事物变化发展的条件，而内因才是事物变化发展的根本，并且外因只能通过内因才能起作用。例如，对大雁来说，食物安全、生命安全、及繁衍安全等是内因，而气候变化只是外因，只有当气候变化威胁到大雁的安全时，才会引起大雁行为的改变。

内因就是指事物的内在矛盾。矛盾是对立双方的统一体，它定义了事物的两种属性（或两个方面）之间的关系，这两个方面既相互斗争又相互统一[17]。例如，冷热就是一个矛盾[2]，而冷和热就是构成这个矛盾的两个方面，它们既相互排斥，又相互依存。矛盾的对立统一才是驱动事物变化发展的动力和源泉。一方面，矛盾存在于任何事物的发展过程之中，事物的发展变化（即事物的行为）都是因为矛盾双方的斗争引起的，没有了矛盾，世界将不再变化。而另一方面，所有事物（即事物的存在）都是矛盾的统一体，它们的性质（即事物的外在表现）都取决于矛盾双方的力量对比[17]。这就像在一个生物体内，其性质都是由内在基因决定的一样，而所有基因都包含一对等位基因，等位基因就像矛盾一样决定着生物体的性质，并控制着生物体的发展变化。

在群体中，个体的存在和发展（即个体的性质和行为）首先取决于其内在矛盾的对立统一，其次又受到其所处环境（包括周围与其有联系的其他个体）的影响，一方面，个体通过与环境交互，从环境中获取资源（包括物质、能量、信息等），并改变个体内部矛盾双方的力量对比，另一方面，群体中所有个体的内部矛盾的综合力量对比会产生某种势能（本文称为群势），这种势能会影响群体内的个体的内部矛盾的发展方向（例如趋同或少数服从多数），第三，个体的行为会改变外部环境、间接影响其他个体的行为，而环境及群体的变化又会对个体行为形成反馈。

与此同时，群体的一切（包括其存在和发展变化）则植根于群体中所有个体的整体表现，当群体表现出时间、空间或功能上可识别的重复发生的有序结构时，就表明群体涌现出了智能[14]。群体的有序结构表现为群体内所有个体的内部矛盾的综合力量对比呈现出某种特定的模式。例如，当越来越多的大雁的安全受到威胁时，它们就会集体迁徙，形成具有全新结构的群体。

因此，我们提出了一种以矛盾为中心的群体智能涌现模型。在该模型中，1) 矛盾决定行为，个体内部矛盾的对立统一是推动个体运动变化的源动力；2) 矛盾反映性质，个体的性质是由个体内部矛盾的双方的力量对比来决定的，而群体的智能特征则是由所有个体内部矛盾的综合力量对比来决定的，群体智能就是个体内部矛盾的特定分布模式的体现；3) 交互影响矛盾，个体与环境之间存在广泛的交互，个体通过交互关联在一起而形成群体，交互同时改变环境及群体，环境反过来影响个体的内部矛盾，群体形成的群势也会影响个体内部矛盾的发展方向。

[2] 因为矛盾总是由两方面构成的，所以有时候也称为“一对矛盾”。

本文的主要贡献如下：首先，我们提出了一种两层的以矛盾为中心的群体智能涌现模型，在内层，个体在内部矛盾的驱动下采取行动，在外层，个体通过交互联系在一起形成群体，交互及群势会影响群体中个体的内部矛盾以及内部矛盾的分布方式，群体智能就表现为内部矛盾的特定分布。我们还形式化地描述了该群体智能涌现模型，给出了个体行为的动力学方程和群体智能的量化公式。其次，根据群体智能模型，我们通过一个实例描述了群体智能的实现过程，并模拟实现了两个系统验证了模型的有效性。

论文后续内容的组织结构如下：第二节描述了群体智能涌现的两层模型，并通过概念模型的方式刻画了模型中的概念（包括个体、群体、环境、群势、群体智能等）的相互关系；第三节形式化地描述了群体智能涌现模型；第四节阐述了群体智能的涌现过程，并通过一个实例展示了群体智能的实现过程；第五节介绍了两个实例系统，通过模拟实验验证了群体智能涌现模型的通用性；第六节与相关工作进行了对比分析；第七节给出了本文工作的总结以及将来的研究方向。

## 2. 以矛盾为中心的群体智能涌现模型概览

在探讨群体智能时，关键在于回答决定个体性质的根基是什么、驱动个体行为的源动力是什么、个体是如何行动的、个体是如何形成群体的、以及群体智能是如何从群体中涌现出来的等问题。我们将群体智能涌现模型分成两层，即个体层和群体层，这两层通过一个反馈环联系在一起（如图1所示）。

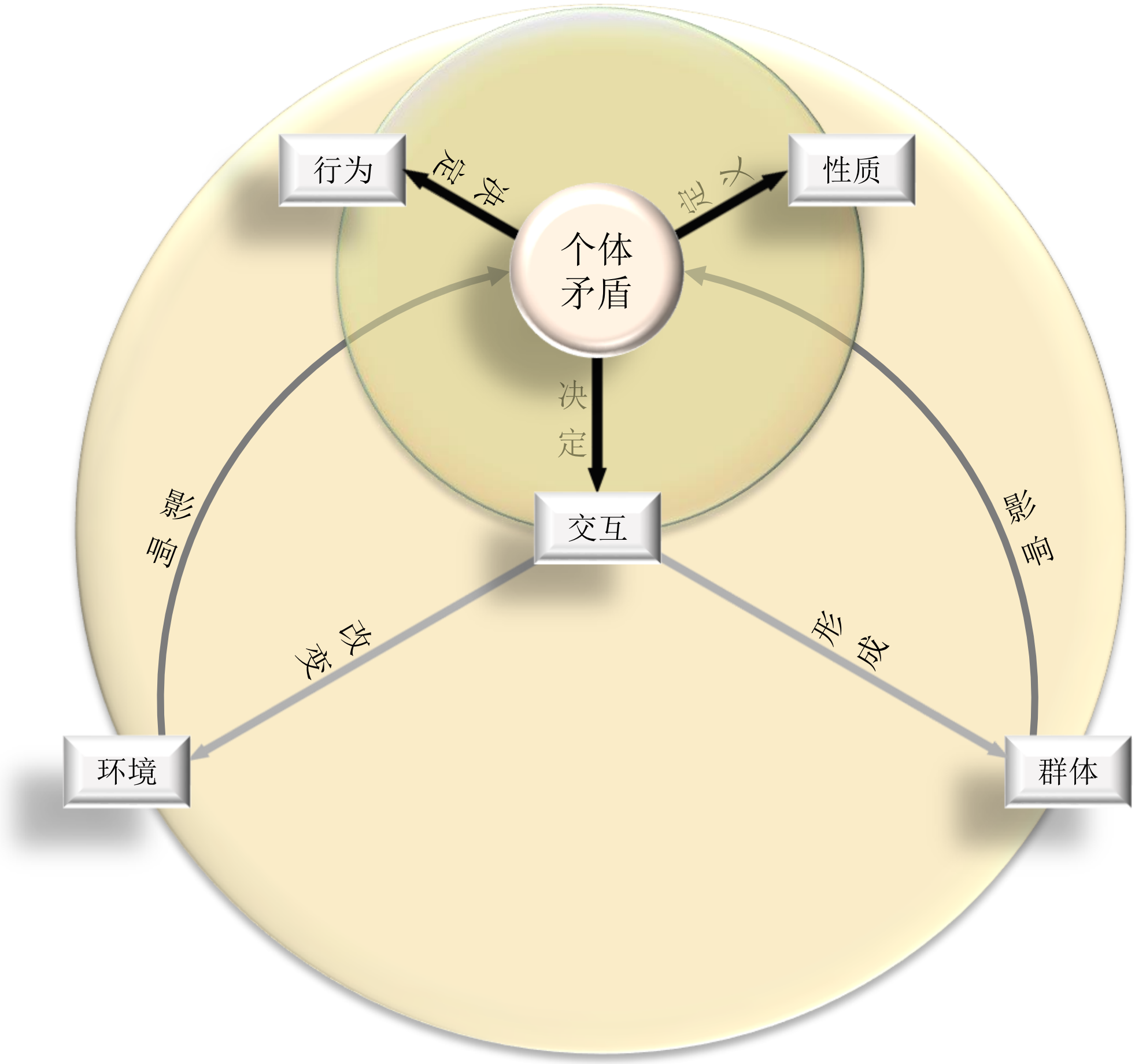

图 1. 群体智能的涌现模型

其中：

- 微观层（或个体层）以矛盾为中心，个体的性质是由个体内部的矛盾定义的，矛盾的运动变化是驱动个体行为的源动力，矛盾的对立/统一（竞争/合作）决定了个体的行为以及个体与环境之间的交互；
- 宏观层（或群体层）以交互为中心，一群个体在与环境交互时，会因为竞争或共享环境资源而联系在一起，处于在环境中的个体相互联系在一起形成群体，群体智能表现为个体集合的整体特征；
- 个体矛盾决定个体之间的交互，矛盾的变化发展的需求决定了个体该如何交互来竞争和占有环境资源。个体通过交互改变外界，外界反过来影响个体的内部矛盾，形成一个始于矛盾、终于矛盾的反馈环：矛盾→交互→环境（包括群体）→矛盾。

在这个模型中，个体和群体的方方面面，包括个体的存在与发展、群体的形成与演化、以及群体智能涌现的条件、动力、方向、途径、及形式等，都是以矛盾为中心展开的（如图2所示）。

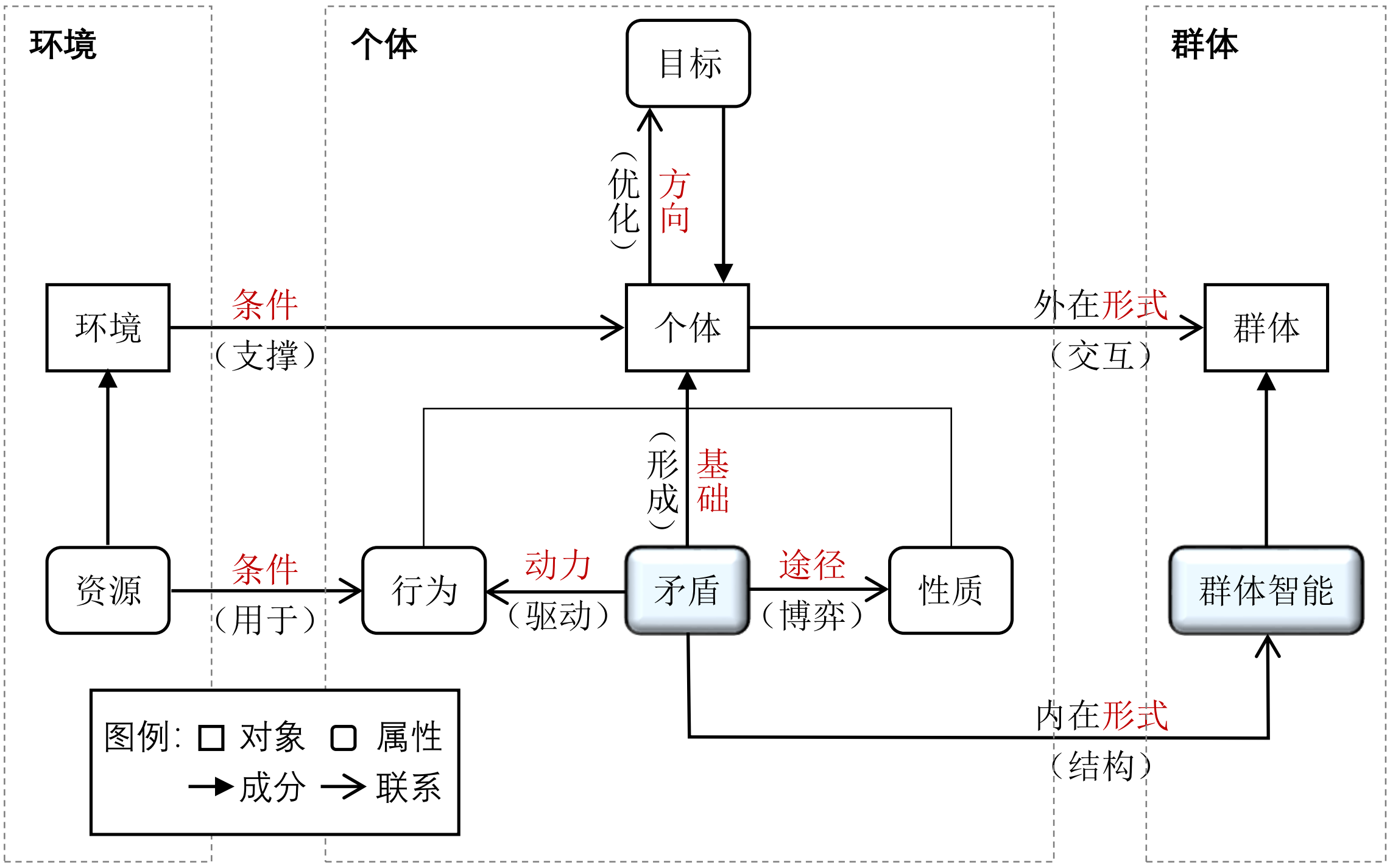

图 2. 群体智能的概念模型

首先（基础），个体的存在是矛盾对立统一的结果。个体就是一系列内部矛盾的统一体，个体区别于其它个体的根本原因在于它具有区别于其它个体的内部矛盾。个体的存在就是矛盾不断斗争的结果（例如一个水库就是进水和排水、扩容和填埋这两对矛盾不断斗争的产物）。矛盾的统一保证了个体性质的相对稳定，维持了个体的存在（例如水库的进水和排水以及扩容和填埋保持相对的稳定才能使水库一直存在）。

其次（动力），矛盾双方的斗争是个体变化发展的源动力。矛盾双方总是企图壮大自身的力量而削弱对方的力量，从而引起矛盾双方力量的此消彼长，乃至矛盾的破裂（一方完全战胜另一方）。矛盾双方力量对比的变化会改变个体的性质，而矛盾的破裂会导致旧矛盾的消失或新矛盾的产生，进而引起个体的演化。个体的目标是为了在不断变化的环境下维持自身的存在（即维持矛盾统一体的相对稳定性），并不断演化（即形成新的矛盾统一体）以适应环境的变化。

第三（形式），群体智能的涌现是个体矛盾不断积累的结果。群体智能表现为群体具有了某种特殊的时间、空间或功能上的有序结构，而所有个体矛盾在群体中必然呈现某种分布，群体智能本质上就是具有特定整体结构的矛盾分布的反映。例如，在觅食的蚁群中，蚂蚁的“探索”（explore）（寻找新的食物源）和“利用”（exploit）（从旧的食物源搬运食物）这一矛盾驱动着蚂蚁尽快找到或获取尽量丰富的食物，当足够多的蚂蚁“利用”已经找到的食物源时，就会呈现出蚁群的最短觅食路径。

第四（条件），外界环境既是个体和群体存在和发展的条件，也是群体智能涌现的条件。个体的生存与发展必然需要从环境中汲取大量的物质、能量或信息，个体只有不断与外界进行交互，才能获取环境资源并维持自身的生存和发展。但外界环境只能通过个体的内部矛盾才能对个体的行为产生影响，交互将影响个体矛盾双方力量的对比以及矛盾在群体中的分布，从而改变个体的行为、以及群体中矛盾的分布结构、并涌现出不同的群体智能。

第五（途径），矛盾双方的斗争本质上是一种博弈，矛盾双方通过竞争或合作来实现个体利益的最大化（即优化目标，获得最有利的生存和发展状态）。个体内部矛盾之间的博弈不仅受到个体自身利益的驱动，也受到外界环境以及周围群体的影响。一方面，个体与环境的交互会改变个体内部矛盾双方的力量对比，另一方面，周围个体的内部矛盾在群体中呈现出的分布状况会产生群势，群势会反过来影响个体的行为取向（例如跟风或叛逆）。

第六（方向），个体之间交互的目的是为了个体尽可能多地占用环境资源，个体内部矛盾之间的博弈则在占有环境资源的基础上追求个体最佳的生存与发展状态。直觉上，当个体不再占用环境资源，并且内部矛盾之间的斗争基本平息（即矛盾双方的力量对比保持持平），个体将进入某种假死状态（即极度平衡或静止），群体也就会陷入混沌，反过来，当个体竞争环境资源并追求自身利益最大化时，无论个体内部矛盾还是个体之间必然出现剧烈的竞争，从而远离平衡态，当大量个体远离平衡态时，就可能呈现出特定的整体结构并涌现出特定的群体智能。例如，在迁徙的鸟群中，当大量的鸟在占据有利位置、充分获取周围鸟群的活动信息等环境资源的情况下，同时又能尽可能地保证飞行安全及飞行距离，这时群体就可能呈现出某种特殊的飞行队形结构，也就是涌现出了群体智能。

## 3. 群体智能涌现模型的形式化

本节将形式化地定义以矛盾为中心的群体智能涌现模型中的各种实体和概念，并严格地描述个体及群体的动态性质，包括个体的生存与发展的动力模型、以及群体智能的涌现机理。

### 3.1. 矛盾

**定义 1. 矛盾**，一个（或一对）矛盾由相互对立的方面（或属性）构成。

$$c = <\varsigma, \bar{\varsigma}> \tag{1}$$

其中，$\varsigma$和$\bar{\varsigma}$分别代表对立的方面，即正方和反方。矛盾的正方和反方是相对的，即$\bar{\bar{\varsigma}} = \varsigma$。

我们用$|\varsigma|$和$|\bar{\varsigma}|$表示矛盾双方的（绝对）力量，$|\varsigma|, |\bar{\varsigma}| > 0$，这表明矛盾双方相互依存，它们必然同时存在，一旦一方的力量为$0$，矛盾则不再存在；用$\|\varsigma\|$和$\|\bar{\varsigma}\|$表示矛盾双方的相对力量，$\|\varsigma\| = \frac{|\varsigma|}{|\varsigma|+|\bar{\varsigma}|}$，$\|\bar{\varsigma}\| = \frac{|\bar{\varsigma}|}{|\varsigma|+|\bar{\varsigma}|}$，$\|\varsigma\| + \|\bar{\varsigma}\| = 1$，这表明矛盾双方又相互斗争，一方

力量的增强必然以另一方相对力量的减弱为代价。

另外，用$\Lambda_c$表示矛盾双方的力量对比，$\Lambda_c = \|\varsigma\| - \|\bar{\varsigma}\|$，$\Lambda_c \in (-1,1)$。它反映了矛盾的尖锐程度，$\Lambda_c$的绝对值越大，表明矛盾越尖锐。当$\Lambda_c$趋近于0时，表明矛盾进入相对平衡的状态。

矛盾双方总是以战胜对方为目的（即壮大己方力量、削弱对方力量），但一旦力量较弱的一方失败（即力量变为0），矛盾将不复存在，这会改变个体的本质。因此，矛盾双方是采取斗争的姿态还是维持相对的平衡（即矛盾双方的博弈）取决于个体当前的性质是否有利于其生存和发展。下文在刻画个体的行为时，将具体描述矛盾双方的博弈是如何决定个体的行为的。

## 3.2. 个体

**定义 2. 个体**，一个个体由一组矛盾定义；个体的一切活动都是围绕这些矛盾展开，个体的活动都是为了增强或减弱矛盾的某一方的力量；个体的性质通过矛盾的状态及其变化表现出来；个体需要占有其赖以生存的环境资源；个体存在的目标就是为了优化其性质，使其在当前环境下最有利于其生存和发展。一个个体可以用一个 6 元组来刻画：

$$\iota = < \Gamma, \succcurlyeq, R, A, P, \mu > \tag{2}$$

其中，

- $\Gamma$是矛盾集合，$\Gamma = \{c_1, c_2, \dots, c_N\}$。
- $\succcurlyeq$是一种偏序关系，$\succcurlyeq \subseteq \Gamma \times \Gamma$，它定义了矛盾在个体中的重要程度，越重要的矛盾对个体的性质的影响力越大，$c_i \succcurlyeq c_j$表示对个体的性质来说，矛盾$c_i$的影响力比$c_j$大。$\succcurlyeq$关系是动态可变的。
- $R$是个体生存所需的环境资源集合。
- $A$是动作集合，对于$\Gamma$中的每一个矛盾，都存在一组（4 个）动作。设$c \in \Gamma$，存在$\{\alpha^+_{c.\varsigma}, \alpha^-_{c.\varsigma}, \alpha^+_{c.\bar{\varsigma}}, \alpha^-_{c.\bar{\varsigma}}\} \subseteq A$，其中$\alpha^{+/-}_{c.x}$表示增强或减弱矛盾$c$的某一方的力量。个体的行为是这些动作的组合。矛盾的运动变化需要占有和使用环境资源，个体的动作可以定义为个体运用环境资源推动矛盾的运动，$\alpha \in A: 2^R \times \Gamma \to \Gamma$。
- $P$是性质集合，每一个性质是矛盾不断斗争和积累的结果或反映。设$\rho \in P$，性质$\rho$的动力学方程可以定义如下：

$$\rho^{t+1} = f(\rho^t, \Gamma^t, \succcurlyeq^t, \Delta\Gamma^t) \tag{3}$$

  其中上标$t$用来表示时刻，$\rho^t, \Gamma^t, \succcurlyeq^t, \Delta\Gamma^t$分别表示时刻$t$时的性质、矛盾、矛盾在个体中的重要地位、和矛盾变化。
- $\mu$是效用函数，用来评估个体在占有环境资源的前提下，性质满足个体生存和发展目标的程度，$\mu: R \times P \to [0,1]$，个体的目标就是最大化其效用。

在下文描述个体的行为时（见3.4小节），将具体刻画个体是如何在矛盾的驱动下执行最有利于其生存和发展的动作的。驱动力的源泉就在于矛盾双方之间的博弈。

**定义 3. 矛盾的内部博弈**，在个体的所有动作中，每个矛盾（如$c \in \Gamma$）的其中一方都分别与两个动作相关联（即$\alpha^+_{c.\varsigma}$和$\alpha^-_{c.\varsigma}$、$\alpha^+_{c.\bar{\varsigma}}$和$\alpha^-_{c.\bar{\varsigma}}$），分别用来增强或减弱己方的力量。矛盾$c$的内部博弈可以表示为：$g_c = < N, \{A_i\}_{i \in N}, \{\upsilon_i\}_{i \in N} >$，其中$N = \{c.\varsigma, c.\bar{\varsigma}\}$是参与者集合，$A_i$是动作集（或策略集），$A_{c.\varsigma} = \{\alpha^+_{c.\varsigma}, \alpha^-_{c.\varsigma}\}$，$A_{c.\bar{\varsigma}} = \{\alpha^+_{c.\bar{\varsigma}}, \alpha^-_{c.\bar{\varsigma}}\}$，$\upsilon_i$是收益函数，$\upsilon_i: \prod_{i \in N} A_i \to R$。

因为矛盾双方的力量对比总是此消彼长的，因此，试图增强自身的力量相对于竞争，而减弱是一种退让，相对于合作。矛盾双方的博弈可以用一个收益矩阵来表示（表1）：

表 1. 内部矛盾博弈矩阵

| $\varsigma$ \ $\bar{\varsigma}$ | $\alpha^{+}_{c.\bar{\varsigma}}$（竞争） | $\alpha^{-}_{c.\bar{\varsigma}}$（合作） |
|---|---|---|
| $\alpha^{+}_{c.\varsigma}$（竞争） | $(\upsilon_{\varsigma_{11}}, \upsilon_{\bar{\varsigma}_{11}})$ | $(\upsilon_{\varsigma_{12}}, \upsilon_{\bar{\varsigma}_{12}})$ |
| $\alpha^{-}_{c.\varsigma}$（合作） | $(\upsilon_{\varsigma_{21}}, \upsilon_{\bar{\varsigma}_{21}})$ | $(\upsilon_{\varsigma_{22}}, \upsilon_{\bar{\varsigma}_{22}})$ |

其中，$\upsilon_{\varsigma_{kl}}, \upsilon_{\bar{\varsigma}_{kl}}$ $(k, l = 1,2)$分别是矛盾正反方的收益。

## 3.3. 群体及群势

**定义 4. 群体**，一个群体由一组相互交互的个体组成。

$$\Omega =< \Sigma, X > \tag{4}$$

其中，

- $\Sigma$是个体集合，$\Sigma = \{\iota_1, \iota_2, \ldots, \iota_M\}$。
- $X$是个体之间的正在发生的交互的集合，$X = \{\chi_1, \chi_2, \ldots, \chi_K\}$。一群个体在与环境交互时因为竞争环境资源而联系在一起。交互可以是两个个体之间的，也可以是多个个体之间的。为了简单起见，我们假定每个交互都有一个中心个体（相当于从中心个体的视角来看待交互），并且围绕该个体的某个内部矛盾展开，其它个体通过环境与中心个体关联在一起，相互竞争该矛盾变化发展所需的资源。

$$\chi =< \iota_\kappa, c_\kappa, I, e > \tag{5}$$

其中，$I \subseteq \Sigma$是该交互所涉及到的个体的集合，$\iota_\kappa \in I$为中心个体，$c_\kappa \in \Gamma$是该交互会影响到的内部矛盾，$e$ 是与该交互有关的外界环境因素（即$I$竞争的环境资源），$e \subseteq \iota_\kappa.R$，$I$中的其它个体都通过 $e$ 与$\iota_\kappa$关联在一起。

在上面的定义中强调个体参与的交互集是动态的，这是因为个体周围的其他个体可能不是固定的，它在竞争环境资源时所面对的交互对象可能是动态变化的。

**定义 5. 群势**，一个矛盾的尖锐程度在群体中的不同个体内会有所不同，矛盾的尖锐程度在群体中的分布会产生一种势能，即群势。

设$c \in \Gamma$，矛盾$c$的尖锐程度$\Lambda_c$在$\Sigma$中所有个体内的取值分别为$\{\lambda_1, \lambda_2, \lambda_3, \ldots \lambda_M\}$，$p(\lambda_k)$为$\Lambda_c = \lambda_k$时的概率，$c$的尖锐程度在$\Sigma$中的分布的数学期望$E_\Omega(c)$（即平均尖锐程度）为：

$$E_\Omega(c) = \sum_{k=1}^{M} \lambda_k p(\lambda_k) \tag{6}$$

其中$M$是群体中个体的数量。相应地，$\Lambda_c$的信息熵$H(\Lambda_c)$为：

$$H(\Lambda_c) = -\sum_{k=1}^{M} p(\lambda_k) \ln\big(p(\lambda_k)\big) \tag{7}$$

相对于矛盾$c$的群势$P_\Omega(c)$定义为$E_\Omega(c)$与$H(\Lambda_c)$的倒数的乘积[3]：

$$P_\Omega(c) = \frac{E_\Omega(c)}{H(\Lambda_c)} \tag{8}$$

$P_\Omega(c)$的取值范围为$[-1,1]$。当$P_\Omega(c) > 0$时，会对个体带来正面影响（相对于矛盾的正方）；反之，当$P_\Omega(c) < 0$时则会对个体带来负面影响（相对于矛盾的正方）。$P_\Omega(c)$的绝对值越大，表明群势对个体的影响力越强。

因为个体并不总是与群体中的所有其他个体都存在交互，它的行为往往只受到与它交互的周围个体的影响，因此，在考虑群势时，相对于某次交互，周围个体形成的群势对个体行为的影响更有意义。类似地，设$\chi_\iota \in X$是以个体$\iota$为中心的交互，$c \in C$，将$E_\Omega(c)$和$P_\Omega(c)$中的$\Sigma$用$\chi_\iota.I$代替，就可以得到该交互中相对于矛盾$c$的平均尖锐程度$E_{\chi_\iota}(c)$和群势$P_{\chi_\iota}(c)$。

**定义 6. 相对群势**，在一个群体中，群体对不同个体的影响力是不同的，我们将群体对特定个体的影响力称为相对群势。

设$\iota$为群体中的某个个体，$\Lambda_c^\iota$是$\iota$的矛盾$c$的尖锐程度，针对个体$\iota$的、相对于矛盾$c$的相对群势$rP_\Omega(\iota,c)$定义为$\Lambda_c^\iota$与$E_\Omega(c)$之间的反差的函数：

$$rP_\Omega(\iota,c) = \frac{|E_\Omega(c) - \Lambda_c^\iota|}{|E_\Omega(c)| + |\Lambda_c^\iota|} \times \frac{1}{P_\Omega(c)} \tag{9}$$

它表明，与周围个体的反衬越明显，周围个体（即群体）对该个体的影响力越大。

交互将改变个体对环境资源的占有状态，进而影响个体对其当前生存状态的评估。个体在交互时会因为群势而改变对环境资源的竞争策略，同样也会改变个体对环境资源的占有。后文在介绍群体智能的涌现机理时会对个体之间的交互、以及交互和群势对个体行为的影响进行详细的描述。

## 3.4. 个体与群体行为模型

群体智能的涌现就是个体在群体中不断采取行动的结果。本小节将首先描述个体在群体中是如何决策其行为的，包括个体的独立行为和个体在参与交互时（即在群体中）的行为，接着将描述群体智能是如何从个体的行为中涌现的。

### 3.4.1. 独立个体行为

当个体独立行动时，它只需要关心其矛盾状态反映出的性质是否最有利于其生存和发展。如前文所述，个体的行为取决于矛盾的内部博弈。但矛盾双方在博弈时，既要追求矛盾双方的局部利益的均衡，又要追求个体（全局）利益的最大化。对于一个矛盾来说，矛盾双方的均衡既是局部利益，也是短期利益，而个体的生存和发展既是全局利益，也是长远利益。矛盾双方在实现局部均衡和追求全局利益最大化之间的折衷决定了个体该如何执行与该矛盾相关的动作。

**定义 7. 个体行为**，个体行为是一组与矛盾相关的动作的组合。在任一时刻，个体行为会涉及到它的所有内在矛盾，并且与一个矛盾相关的动作总是成对出现的。令$c_i \in \Gamma$（$i = 1..N$），个体的行为集为：

---

[3] 类似于重力势能的公式$P=mgh$，我们用矛盾在群体中的概率分布的数学期望表示矛盾的“重量”，矛盾所蕴含的信息量（或矛盾的不确定程度，即熵）的倒数表示矛盾的“高度”。采用信息熵的倒数的原因在于熵越大表明矛盾分布的不确定度越大、越混乱，即矛盾分布得越不整齐，越难以产生整体的影响力。

$$B = \prod_{i=1..N} \{< \alpha^{+}_{c_i.\varsigma}, \alpha^{+}_{c_i.\bar{\varsigma}} >, < \alpha^{+}_{c_i.\varsigma}, \alpha^{-}_{c_i.\bar{\varsigma}} >, < \alpha^{-}_{c_i.\varsigma}, \alpha^{+}_{c_i.\bar{\varsigma}} >, < \alpha^{-}_{c_i.\varsigma}, \alpha^{-}_{c_i.\bar{\varsigma}} >\} \quad (10)$$

假设个体所有矛盾$\Gamma$的内部博弈为：$g_{c_1}, g_{c_2}, \dots, g_{c_N}$，个体在行为时，总是在维持均衡的前提下，期望实现个体利益的最大化。

$$\begin{aligned} & \max_{b \in B}(\mu) \\ s.t. & \\ & for\ all\ g_{c_i}\ (i = 1..N)\text{: } the\ \text{Equilibrium of}\ g_{c_i}\ achieved \end{aligned} \quad (11)$$

矛盾双方在博弈时的收益并不总是一层不变的，会因为个体状态（包括个体所需的环境资源的状态）的变化而动态变化，也就是说，矛盾双方的博弈在引起个体状态发生改变的同时，个体状态的改变又会反过来影响矛盾双方的收益，矛盾双方的博弈会随着收益的不断变化而形成动态的均衡，这种动态的均衡应该不断促进个体利益的最大化。

### 3.4.2. 群体中个体行为

当个体处于群体之中时，它在决定自己的行为时还需要考虑与其它个体之间的交互及其所处群体的群势带来的影响。

个体一方面通过交互从环境中获取资源，另一方面通过交互与其它个体竞争资源。一般来说，个体占有的资源越丰富，其生存状态越好，而交互会改变个体对环境资源的占有，进而影响个体的生存状态。因此，个体在交互中，总是尽量争取占有更多的资源，以保证自身利益的最大化。但环境中的资源总是有限的，个体之间的竞争必然导致某些个体占有更多的资源，而其它个体则占有较小的资源，所以，个体在决定自己的行为时，要权衡对资源的占有以及自身的生存和发展，在占有更多资源和利益最大化之间取得某种折衷。

设以个体$\iota \in \Sigma$为中心的交互集为$X_\iota = \{\chi_1, \chi_2, \dots, \chi_L\}$，$e_\iota$是个体$\iota$在交互中竞争的环境资源集合$e_\iota = \bigcup_{\chi \in X_\iota} \chi.e$。同样假设个体所有矛盾的内部博弈为：$g_{c_1}, g_{c_2}, \dots, g_{c_N}$，个体在采取行动时，将在维持博弈均衡的前提下，占有尽量多的环境资源，并实现个体利益的最大化。

$$\begin{aligned} & \max_{b \in B}(\mu) \\ s.t. & \\ & for\ all g_{c_i}\ (i = 1..N)\text{: } the\ \text{Equilibrium of}\ g_{c_i}\ achieved, \\ & for\ all\ \epsilon \in e_\iota\text{: } |\epsilon| \le \epsilon_{max}\ and\ |\epsilon| \to \epsilon_{max} \end{aligned} \quad (12)$$

其中，$|\epsilon|$表示个体占用环境资源$\epsilon$的数量，$\epsilon_{max}$表示环境资源$\epsilon$的最大量。

另外，交互中的群体也具有群势，参与交互的个体在竞争和占有资源时也会受到群势的影响。在群势的影响下，个体跟随大多数个体采取行动本质上是为了自身的状态与大多数个体保持一致，也就是使自身矛盾的尖锐程度与大多数个体的平均尖锐程度趋同。直觉上，当个体跟随大多数采取行动时，会形成一股合力并排挤少数，更容易获得资源；但另一方面，因为环境资源是有限的，当资源不足时，跟随大多数采取行动又可能会导致资源竞争更加激烈（即产生内卷），更难获取资源。因此，个体同样需要权衡在群体中顺应群势是否有利于更好地占有资源。

个体在博弈过程中采取行动时，将在维持均衡的前提下，权衡群势对占用资源带来的影响，占有尽量多的环境资源，并实现个体利益的最大化。

$$
\begin{aligned}
&\max_{b\in B}(\mu)\\
&s.t.\\
&for\ \ all\ \ g_{c_i}\ \ (i=1..N):\ the\ \text{Equilibrium of}\ g_{c_i}\ achieved,\\
&for\ \ all\ \ \epsilon\in e_\iota:\ |\epsilon|\le\epsilon_{max}\ \ and\ \ |\epsilon|\to\epsilon_{max},\\
&for\ \ all\ \ \chi_i\in X_\iota\ \ (i=1..L):\begin{cases} rP_{\chi_i}(\iota,\chi_i.c_\kappa)\to 0 & if\ \sum_{\epsilon\in e_\iota}\frac{|\epsilon|}{\epsilon_{max}}<\theta\\ rP_{\chi_i}(\iota,\chi_i.c_\kappa)\to 1 & otherwise\end{cases}
\end{aligned}
\tag{13}
$$

其中，$\theta$是资源竞争强度的阈值，当大于该值时，表示资源已严重不足；$rP_{\chi_i}(\iota,\chi_i.c_\kappa)$是$\chi_i$中的群体对于$\iota$及其矛盾$\chi_i.c_\kappa$的相对群势。$rP_{\chi_i}(\iota,\chi_i.c_\kappa)\to 0$表示$\iota$在矛盾$\chi_i.c_\kappa$方面与$\chi_i$中的群体趋同，$rP_{\chi_i}(\iota,\chi_i.c_\kappa)\to 1$则表示$\iota$与群体背离。

## 3.5. 群体智能

**定义 8. 群体智能**，当群体在时间、空间、或功能上呈现（或涌现）出有序结构时，就表明群体具有了某种智能。

群体的有序结构本质上是个体性质、行为的有序分布的体现，如特定的行为序列或组合、资源占有和分配方式、结果的积累模式等，而个体的性质和行为都是由其内部矛盾决定的，因此，群体智能就是群体中所有个体的内部矛盾的有序分布的体现。

我们在考察或认定群体智能时，并不一定会涉及到群体的方方面面。因为关注点不同，同样的群体在有的人的眼里可能是无序的，而在另外一些人的眼里却是有序的，反之亦然。例如，在一个可以随意发言的教室里，有人会觉得乱哄哄的，没有组织纪律，但有人却觉得研讨氛围活跃，是一种很好的学习状态。出现不同判断的原因在于他们关注的矛盾不同，只要他们所关注的矛盾是有序的，他们就会认为群体是有序的。因此，在定义群体的智能时，我们要考虑所关注的矛盾集合的大小。

首先，群体内任意一种矛盾的有序分布都可以看成是群体的一种微观（或局部）智能（即相对于矛盾$c$的群体智能$SI(c)$）。设$c\in\Gamma$，$p(\lambda_k)$为$\Lambda_c=\lambda_k$的概率，我们将矛盾$c$的有序度$O(c)$定义成$\Lambda_c$的信息熵$H(\Lambda_c)$的函数（借鉴[25]中的思想，它用信息熵来量化群体智能的涌现），其中$M$是群体中个体的数量，$\ln(M)$是$H(\Lambda_c)$的最大值。

$$SI(c)=O(c)=\frac{\ln(M)-H(\Lambda_c)}{\ln(M)} \tag{14}$$

其次，一组（或全部）矛盾的有序分布可以看成是群体的一种宏观（或全局）智能（即相对于矛盾集$\Gamma$的群体智能$SI(\Gamma)$），这组（或全部）矛盾的有序度$O(\Gamma)$可以定义成它们的尖锐程度的联合熵$H(\Gamma)$的函数，其中$M$是群体中个体的数量，$N$是矛盾的个数。

$$H(\Gamma)=-\sum_{\Lambda_{c_N}}\dots\sum_{\Lambda_{c_1}}p\left(\Lambda_{c_1},\dots,\Lambda_{c_N}\right)\ln\left(p\left(\Lambda_{c_1},\dots,\Lambda_{c_N}\right)\right) \tag{15}$$

$$SI(\Gamma)=O(\Gamma)=\frac{\ln(M^N)-H(\Gamma)}{\ln(M^N)} \tag{16}$$

从上面的定义中可以看出：

1. 群体智能有高有低、有大有小。熵$H$越小，有序度$O$越大，表明矛盾在群体中的分布越有序，群体智能就越高。关注的矛盾集越大，群体智能越容易观察到，表明群体

智能越具有宏观性，群体智能越大。

2. 关注的矛盾集不同，对群体智能的判断（或认知）也会不同。例如，在一个教室里，如果关注的是与自由讨论相关的矛盾，讨论得越热烈、交流对象越随意则说明群体越有序、群体智能越高；反之，如果关注的是听课秩序，大声讲话或随意交谈则意味着秩序混乱，群体智能不高。

设任意两个时刻群体相对于矛盾$c$的群势及智能分别为$P_{\Omega}^{1}(c)$、$P_{\Omega}^{2}(c)$、$SI^{1}(c)$和$SI^{2}(c)$，可以看出当矛盾$c$的尖锐程度在群体中的概率分布的数学期望保持不变时，有：

$$\frac{\Delta|P_{\Omega}(c)|}{\Delta SI(c)}=\frac{\left|P_{\Omega}^{2}(c)\right|-\left|P_{\Omega}^{1}(c)\right|}{SI^{2}(c)-SI^{1}(c)}=\frac{|E_{\Omega}(c)|\ln(M)}{H^{1}(\Lambda_{c})H^{2}(\Lambda_{c})}>0 \quad (17)$$

因此，我们有如下的断言：

**断言 1.** 群体智能越高，群势越强（或越大）；反之亦然。

# 4. 群体智能的涌现与实现

## 4.1. 群体智能的涌现过程

根据前面的描述，我们可以看出群体智能的涌现是在两个层面上同步展开的。

首先，在个体层，个体的行为完全受其内在矛盾的驱动，个体根据内在矛盾之间的博弈决定自身的行为，以实现个体利益的最大化。个体的行为会受到所处环境的资源的约束以及所在交互群体的群势的影响，环境和群体的约束会影响个体获取生存资源的难度和成本，个体会不断调整内部矛盾之间的博弈，进一步影响自身行为的决策过程。

其次，在群体层，主要是指与个体存在交互的（子）群体，在群势的作用下，一方面，具有相近矛盾尖锐程度的个体会形成一个簇，同化周边的其它个体，即影响其它个体的内在矛盾的状态与这个簇保持一致，从而壮大这个簇的规模，另一方面，个体会受到周边强势簇的影响或吸引，调整自身内在矛盾的状态，融合到簇中从而获得更佳的资源竞争优势。强势的簇就是一个局部有序的子群体。

第三，个体在不同的矛盾驱动下会与不同的交互对象竞争环境资源，另外，个体的移动也会改变交互对象，这会导致新簇的形成或已有簇的解散或扩张。

第四，群体可能会存在多个簇，不同的簇可能因为影响范围的扩大而产生交叉，进而引起簇的相互侵蚀或融合。

第五，当某个（或某些）簇已经在全局范围产生影响时，就表明群体涌现出了某种群智。

## 4.2. 群体智能的实现

根据前文描述的群体智能涌现模型以及涌现过程，要实现一个群体，首先要定义个体及其交互，接着实现个体。考虑到环境资源是个体赖以存在的前提条件，而竞争环境资源又是个体之间产生交互的根本原因，所以，在实现个体及其交互时，还需要先定义环境资源。

为了便于描述和理解群体智能的实现过程，先描述一个例子，后面的描述都围绕这个例子展开。例如，在觅食的蚁群中，蚂蚁都在寻找食物或搬运食物；一旦找到食物源，它们总是试图尽快将食物搬运到巢穴；卸下食物后，则试图快速回到已找到的食物源处；在没有搬运食物时，它们也会寻找新的食物源。

在下文的具体描述中，将根据前文提出的概念模型来描述群体的各种成分。

### 4.2.1. 环境资源

在蚁群所处的环境中，有蚂蚁的活动范围（即觅食空间），它以蚁穴为中心，其上分布着若干食物源，食物源的位置对蚂蚁来说可能总是未知的（假设蚂蚁没有长期的记忆能力），蚂蚁为了便于记录食物源到蚁穴之间的路径，会在路径上分泌信息素，这些信息素除了自己能感知到之外，也会被其它蚂蚁感知到。

环境中的资源可以用资源的类型（物质、能量、或信息）、资源的数量、资源是共享型的还是独占型的、以及资源是消耗性的还是再生性（可重用）的等几个方面的来刻画。蚁群所处环境中的资源列表如下（表 2）：

表 2. 蚁群环境资源列表

| 资源名称 | 类型 | 数量 | 共享/独占 | 消耗/再生 |
|---|---|---|---|---|
| 觅食空间 | 物质 | $N\times N$ | 独占 | 再生 |
| 蚁穴 | 物质 | 1 | 共享 | 再生 |
| 食物 | 物质 | $\leq M\times K$ | 独占 | 消耗 |
| 信息素 | 信息 | 不定 | 共享 | 消耗 |

其中，觅食空间是一个 $N\times N$ 的矩阵空间，矩阵的每格内在一个时刻只能有一只蚂蚁；空间内有 $M$ 个食物源，每个食物源的食物总量不超过 $K$（即 $K$ 只蚂蚁的搬运量），每份食物只能由一只蚂蚁搬运；蚂蚁分泌的信息素的总量是动态变化的，信息素分布在觅食空间的网格中，一个网格内的信息素可以不断叠加，同时信息素会不断挥发直至完全消失。

### 4.2.2. 个体

根据前文对群体涌现模型的描述以及个体的定义，在实现个体时，需要先刻画个体的外在表现（包括行为和性质），再由表及里分析个体的内在构成（即矛盾及矛盾在决定个体性质时的重要程度），最后评估个体目标的实现状况（即效用函数）。

**目标**。在群体中，个体的目标都是为了更好地生存和发展。具体到蚁群中，蚂蚁的目标是找到丰富的食物（源），并尽快将食物搬运到蚁穴中；另外，一旦找到食物源，它总是期望以后依然能快速找到该食物源。

**性质**。个体的性质（即外在表现）是个体可观察属性的综合体现。在蚁群中，每一只蚂蚁都在觅食空间内不断游走，会呈现出特定的游走路线。蚂蚁是否搬运着食物、蚂蚁所处的位置、以及蚂蚁的游走方向是可观察到的一些属性，蚂蚁的性质（即游走路线）就是从这些属性中展现出来的。蚂蚁是否能更好地实现目标，取决于当它处于空载状态时，是否正行走在能快速找到食物源的路上，而当它处于负载状态时，是否正行走在能快速回到蚁穴的路上。假设蚂蚁在觅食空间中已经找到了 $M$（≥0）个食物源（$FS_M$），根据蚂蚁当前的游走路线偏离蚁穴到食物源之间的最短路径的程度，可以评估蚂蚁的效用（见定义 2）。

$$\mu=\max_{fs\in FS_M}\left(sim\big(\mathrm{Route}(self),\mathrm{Path}(nest,fs)\big)\right) \tag{18}$$

其中，$sim(.,.)$为一个相似度函数，它计算蚂蚁最近一次从蚁穴出发（或返回蚁穴）的游走路线 Route(*self*)和从蚁穴到某个食物源的最短路径 Path(*nest*, *fs*)之间的相近程度。

**矛盾**。蚂蚁的性质都是由其内在矛盾的状态决定的。通过分析决定蚂蚁的游走路线，可以发现决定蚂蚁游走路线的内在矛盾包括：是探索新的食物源，还是利用旧的食物源；是否能安全避开其它蚂蚁或者存在与其它蚂蚁发生碰撞的危险（见表 3）。

**表 3. 蚂蚁的内在矛盾**

| 矛盾 | 描述 |
|---|---|
| $c_1$. *探索* 或 *利用* | 当觅食（没有搬运食物）时，是寻找新的食物源，还是重新回到已找到的食物源；<br>当正在搬运食物时，是探索新的返回蚁穴的路径，还是沿着其它蚂蚁走过的路径 |
| $c_2$. *安全* 或 *碰撞* | 是否会与周围的其它蚂蚁发生碰撞 |

蚂蚁的性质与内在矛盾之间的关系如表 4 中所示，其中也显示了在不同情况下，内在矛盾的重要程度，即对蚂蚁性质的影响的大小。矛盾的重要程度表明，当蚂蚁搬运食物时，安全回到蚁穴是最重要的；而当蚂蚁外出觅食时，如果已经知道存在食物源，游走安全是最重要的，如果尚不知道食物源，则找到新的食物源更为重要。

**表 4. 蚂蚁的性质与内在矛盾之间的关系**

| 性质 | 内在矛盾（及重要程度） |
|---|---|
| 搬运食物时的游走路线 | $c_2 \succcurlyeq c_1$ |
| 回到已知食物源时的游走路线 | $c_2 \succcurlyeq c_1$ |
| 寻找新的食物源时的游走路线 | $c_1 \succcurlyeq c_2$ |

蚂蚁正是在当前矛盾的驱动下，不断形成新的游走路线。例如，在蚂蚁觅食时，随着$c_1$的双方力量对比$\Lambda_{c_1}$的变化，蚂蚁不断靠近或偏离其它蚂蚁留下的游走路线，当$\Lambda_{c_1}$趋近于$-1$时（假设“利用”为反方），表明蚂蚁总是走在通往已知食物源的路径上。

**行为（动作集）**。蚂蚁的动作就是改变其内在矛盾双方的力量对比，与矛盾相关的动作集见表 5.

**表 5. 蚂蚁的内在矛盾及相关的动作**

| 矛盾 | 动作集 |
|---|---|
| $c_1$ | 觅食时，*随机游走*，以探索新的食物源；或者感知环境中遗留的信息素，并根据信息素的浓度决定*随机移动*到相邻的位置、或*循迹移动*到信息素浓度高的位置<br>搬运食物时，释放信息素；*随机游走*，以探索新的返回路径；或者感知信息素，并*循迹移动*并返回蚁穴 |
| $c_2$ | 根据周围蚂蚁的数量选择*进入*或*远离*相邻位置（相邻位置周围的蚂蚁越多，表明发生碰撞的可能性越高） |

**矛盾的内部博弈**。个体如何选择动作是由内部矛盾的双方之间的博弈决定的。例如，当蚂蚁觅食时，对于“探索或利用”这一矛盾，蚂蚁如何行动（包括移动方式以及移动到什么样的位置）由如下的博弈矩阵决定（表6）。

**表 6. 蚂蚁觅食时矛盾 $c_1$（探索/利用）的博弈矩阵**

| 利用 / 探索 | 竞争（往高浓度处） | 合作（往低浓度处） |
|---|---|---|
| 竞争（随机游走） | $(\Delta P_{new_r}, \Delta P_{old_h})$ | $(\Delta P_{new_r}, \Delta P_{old_l})$ |
| 合作（循迹行走） | $(\Delta P_{new_f}, \Delta P_{old_h})$ | $(\Delta P_{new_f}, \Delta P_{old_l})$ |

因为蚂蚁会根据信息素浓度的高低来判断找到旧的食物源的可性能，一般来说，较高的信息素浓度意味着循着遗留有信息素的路径可以找到食物源，但过低的信息素浓度意味着旧的食物源可能已经被搬空，因此，蚂蚁将要移动到的位置处的信息素浓度的高低、以及蚂蚁远离或靠近遗留有信息素的位置都会影响蚂蚁对找到新的（或旧的）食物源的判断，所以，在“探索或利用”这对矛盾的博弈矩阵中，可以将矛盾双方的收益定义为蚂蚁进入相邻位置后找到新的（或旧的）食物源的概率的变化，其中，$\Delta P_{new_r}$和$\Delta P_{new_f}$分别表示随机或循迹移动后找到新的食物源的概率的变化，$\Delta P_{old_h}$和$\Delta P_{old_l}$则分别表示往信息素高（或低）的位置移动后回

到旧的食物源处的概率的变化。这样，随着矛盾双方的博弈，蚂蚁就会在“探索新的食物源”和“利用旧的食物源”之间达成某种平衡。

### 4.2.3. 群体（个体交互）

个体在采取行为的过程中，需要占用环境资源，当个体之间竞争或共享环境资源时，个体之间就会发生交互。在蚁群的环境中，觅食空间内的位置是独占型资源，信息素是共享型资源，因此，当蚂蚁移动时会因为竞争位置资源而产生（直接）交互，同时，也会因为在环境中释放信息素或感知环境中遗留的信息素而产生（间接）交互（表 7）。

表 7. 蚂蚁之间的交互

| 交互 | 影响到的主要矛盾 | 交互个体集合 | 竞争/共享的资源 |
|---|---|---|---|
| 移动 | 安全/碰撞 | 周围蚂蚁 | 位置 |
| 释放/感知信息素 | 探索/利用 | 所有空闲<br>或 忙碌的蚂蚁 | 信息素 |

当蚂蚁移动时，如果周围的蚂蚁比较多，对位置的竞争会比较激烈，这时，相对于“安全/碰撞”这一矛盾而言，“碰撞”一方的力量大于“安全”一方，周围蚂蚁的群势会趋向于-1（假设“碰撞”一方是反方），因此，蚂蚁在交互中将选择背离群体，即寻找更“安全”的位置移动。当蚂蚁感知到信息素时，因为信息素是共享资源，不会导致资源紧张，所以蚂蚁总是采取与其它蚂蚁相似的行为寻找食物源或返回蚁穴。

### 4.2.4. 智能涌现

根据前面的描述，在没有找到食物源之前，蚂蚁会在觅食空间内随机游走。在找到食物源之后，蚂蚁会搬起食物并往蚁穴移动，在移动过程中会沿途遗留信息素。此后，蚂蚁在游走过程中，感知到信息素时，在保证游走安全的情况下，会根据相邻位置的信息素浓度的高低、以及周围蚂蚁的行为方式，跟随其它蚂蚁往返于蚁穴和食物源之间。

在蚂蚁刚刚找到食物源时，信息素还没有散布开，路过的蚂蚁在感知到信息素之后会聚集在信息素浓度较高的通往食物源的路径上，随着信息素不断从食物源向着蚁穴扩散，聚集的蚂蚁形成的游走路线会不断从食物源延伸到蚁穴，从而形成一条从蚁穴到食物源的觅食路径。

由于蚂蚁在返回蚁穴的过程中，为了避免与其它蚂蚁发生碰撞，并不总是沿着从食物源到蚁穴的直线路径游走，这样就可能形成多条觅食路径。但因为蚂蚁在没有碰撞危险时总是优先选择能最快返回蚁穴的路径，于是这条路径上的信息素的浓度通常会高于其它路径，从而形成一条从食物源到蚁穴的最佳返回路径，反过来也就成了蚂蚁的最佳觅食路径。

如前文所述，蚁群相对于矛盾集（探索/利用、安全/碰撞）的群体智能表现为这些矛盾在蚁群中的分布，即如果已找到有足够食物的食物源，越来越多的蚂蚁就会偏向“利用”和“安全”，导致内部矛盾的尖锐程度的联合熵越来越小，进而展现出越来越高的群体智能。

## 5. 实例研究

为了验证所提出的群体智能涌现模型的可行性和通用性，我们实现了两个模拟系统，这些系统中有的会产生相对稳定的结构或组织模式，有的则会出现相对统一的行为模式。通过这些模拟系统，可以看出个体在矛盾的驱动下以及群势的影响下，在不断的交互过程中能逐

渐涌现出群体智能，同时也表明该模型可用来刻画不同类型的群体智能的涌现。

## 5.1. 大雁迁徙队形的涌现

大雁在迁徙过程中（主要指飞行时），需要考虑两方面的因素，一是它自身的飞行状态，另一个是它在大雁群中的状态。决定大雁行为的内部矛盾主要包括飞行的安全与省力、以及与群体的疏远与贴近。对它自身来说，它可以通过跟随其它大雁来节省力气，但同时要避免跟得太紧而发生碰撞；相对于雁群来说，它要避免远离雁群而掉队，但也不能因为过于拥簇而影响其自由飞翔。因此，对雁群中的大雁来说，保持其飞行既安全又省力、同时与群体若即若离是最有利的。

在模拟系统中，随机产生了若干（10~20 只）的大雁，其中处于最前头的自动被看作是领头雁，其它大雁则在领头雁的带领下迁徙。跟随飞行的大雁总是尽可能地维持其内部矛盾的平衡，以追求其利益的最大化。相对于飞行的安全与省力来说，大雁通过加速或减速来维持它和身前的大雁之间的飞行距离；相对于群体的疏远与贴近来说，大雁通过左右移动（飞远或飞近）来维持它与领头雁的距离及在雁群中的位置。

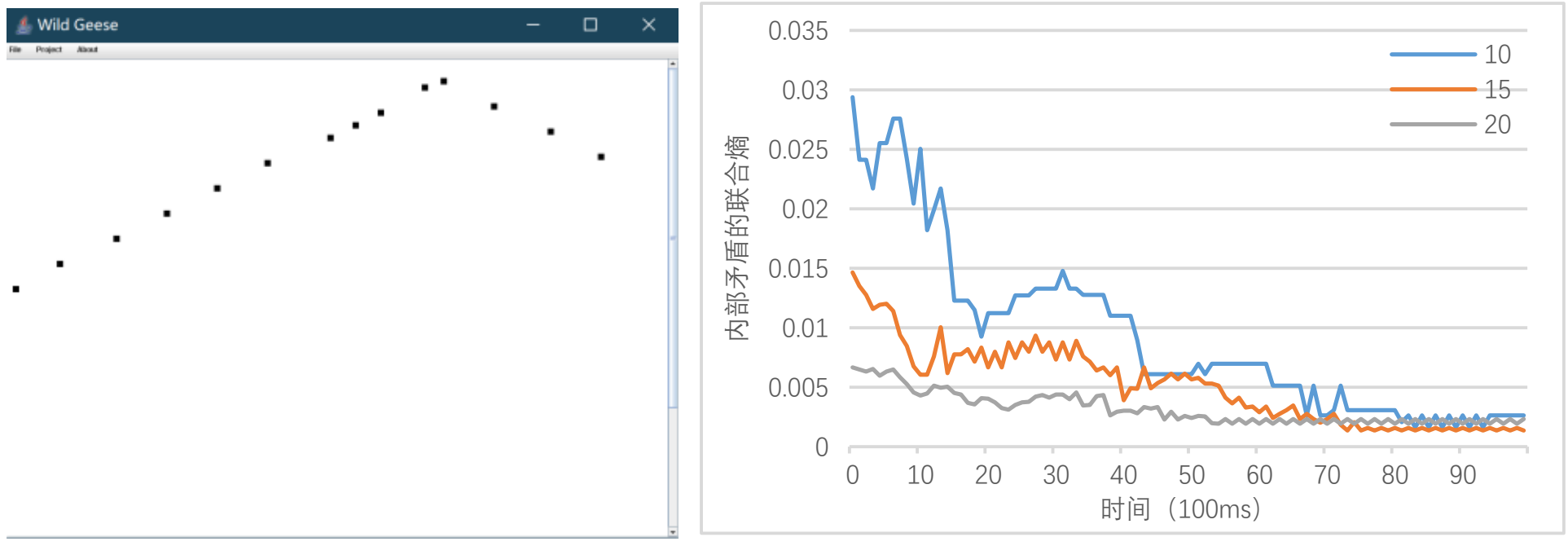

图 3. 大雁的飞行队形及内部矛盾的联合熵

图中（图 3）左侧显示了雁群在飞行一段时间后行程的队列，右侧则显示了不同大小的雁群在飞行过程中大雁内部矛盾的联合熵。从中可以看出，随着时间的推移，雁群内部矛盾的熵越来越小，并趋于稳定，这和雁群形成了相对稳定的飞行队列这一现象相吻合，这表明雁群中涌现出来某种群体智能（即形成了特定的组织模式）。

## 5.2. 囚徒困境与合作的涌现

在研究囚徒困境时，人们总是根据博弈来推理囚徒的行为决策，然后利用重复博弈来模拟合作的出现。在博弈中，总是假设参与博弈的个体都是理性的，能综合推理和计算不同行为策略下的收益；在重复博弈中，则会设定参与者如何根据之前的博弈结果来调整当前的博弈策略的一些原则，例如“以德报怨”（总是合作）、“以牙还牙”（合作对合作、背叛对背叛）等。

但是，在现实世界中，参与博弈的人并不一定都具有严谨的理性分析和推理能力，也不一定会在每次博弈时都进行复杂的数学分析和推理。其次，博弈的对象并不是一层不变的，甚至很多时候博弈双方都是一次性博弈。因此，现有的关于合作的涌现的模拟和研究并不是现实世界中合作的涌现过程的真实写照，并没有挖掘出合作的涌现的真正根源。

其实，现实世界中的人们大多数时候都是凭着直觉在进行决策并采取行动。一方面，他们凭着经验来判断接下来要采取的行为可能会带来的好处或风险，另一方面，他们会参考社

会氛围（即周围人群的普遍行为方式）来调整自己的行为。

在我们实现的模拟系统中，个体内部只有一个矛盾，即合作意愿与背叛意愿，很显然，合作的意愿越强烈，越倾向于采取合作的行为。另外，个体随机分布并游走在一个区域内，当遇到其他人时，会相互传递信息，这些信息主要是关于合作态度的。

个体的合作（或背叛）意愿主要受到三个因素的影响。一是行为的累积收益，即累计每次行为产生的收益。二是后悔的预期累积收益，即累计假设采取相反的行为可能带来的收益。如果合作的累积收益与背叛的后悔累积收益之和较大，则提高自己的合作意愿，反之则降低合作意愿。三是周围个体的行为倾向，如果周围大多数人有较强的合作（或背叛）意愿，则提高合作（或背叛）意愿。

$$\Delta I_c = \begin{cases} 1 & g_c + l_{\bar{d}} > g_d + l_{\bar{c}} \ \ or \ \ |C_{neigh}| > |D_{neigh}| \\ -1 & otherwise \end{cases} \qquad (18)$$

其中，$I_c$表示合作意愿的强弱，$\Delta I_c$为每次交互后合作意愿的变化，$g_c$和$g_d$分别表示合作和背叛的累积收益，$l_{\bar{c}}$和$l_{\bar{d}}$则分别表示反悔合作和反悔背叛行为可能产生的累积收益，$|C_{neigh}|$和$|D_{neigh}|$分别表示周围个体中有合作意愿和没有合作意愿的个体的数量。

在模拟系统中，个体随机分布在一个 100×100 的网格中。为了更好地模拟现实世界中的交互，我们设置了两种场景，一是个体不移动，只与他周围相邻的个体交互，二是个体可以随机移动，在移动的过程中，与所处位置的相邻个体交互。在两个个体交互时，他们选择招供或拒供所获得的收益如下矩阵所示（表 8）。

**表 8. 个体交互的收益矩阵**

| 个体 B ╲ 个体 A | 招供 | 拒供 |
|---|---|---|
| 招供 | (1， 1) | (5， 0) |
| 拒供 | (0， 5) | (3， 3) |

我们分别统计了在个体不移动或移动的情况下，不同群体大小的人群中涌现出合作的概率。如图 4 所示，展示了个体之间进行 100 轮博弈的过程中，规模分别为 1000 至 5000 的群体中有合作意愿的个体所占的比例。

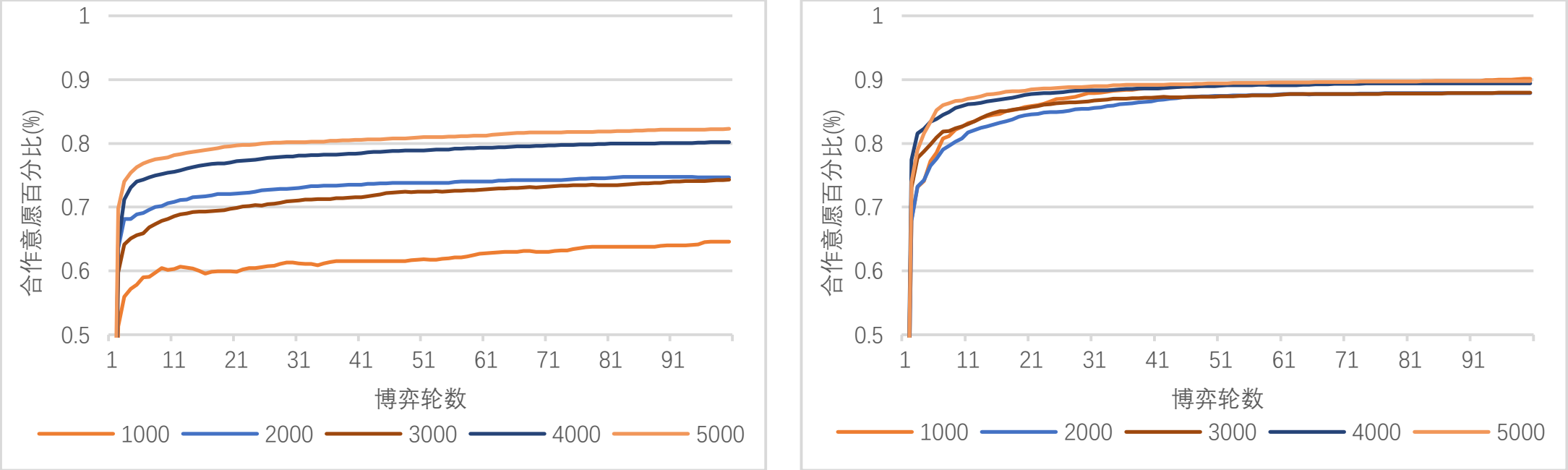

a. 个体不移动　　　　b. 个体移动

**图 4. 重复博弈中合作意图的涌现**

从图中可以看出，随着博弈的轮数增加，群体中有合作意愿的个体比例快速增加并达到某种稳定状态。群体的规模越大，群体中会越快涌现出合作意愿，并且具有合作意愿的个体的占比会越高，这展现了群体智能的规模可伸缩性[3]。当个体可以在群体中自由移动时，具有合作意愿的个体所占的比例会更高。这说明，个体能够接触并交互的对象越多，个体之间的交互越频繁，合作意愿的涌现会越快、在群体中的普遍程度也越高。这与我们在现实世界中观察到的合作的涌现现象非常吻合，随着人口的增加以及人口的迁徙，合作会更快地涌现出来，合作也会更加普遍。

## 6. 相关工作

在群体智能研究中，有的期望建立通用的群体智能涌现理论，为一切涌现现象提供解释，而大多数研究则基于人们对特殊现象（包括自然的、社会的、物理的等）的观察和启示来建立面向特定复杂问题的解决方案。

### 6. 1. 通用的群体涌现理论

群体智能的涌现现象在自然界、社会和人工系统中广泛存在，为此，有越来越多的研究关注群体智能的涌现原理及其建模。因为群体智能常常和自组织、集体等概念联系在一起[8]，涌现被认为是（多主体）系统呈现出高智能特性的过程[19]，具有高智能特性的群体往往呈现出空间、时间或功能上的有序结构，而自组织正是群体朝向有序结构自主演化的过程[12]，因此与群体智能涌现相关的理论成果大多出现在自组织系统领域。其中最具有影响力的成果包括 Prigogine 的耗散结构理论[23]、Haken 的协同学[13]、Thom 的突变论（或灾变论）[4]、Eigen 的超循环论[7]、Mandelbrot 的分形理论[22]、以及 Lorenz 的混沌理论等[21]。在此基础上也形成了一些复杂、动态、多层次的群体理论[1][28][30]。

但是，这些理论一般只考虑或研究了群体智能涌现的某一（些）方面。例如耗散结构理论主要探讨了有序结构涌现的外部环境条件，认为不断从外部环境交换能量是系统维持结构稳定有序的**必要条件**。协同学提出系统演化的**动力**（微观上）来源于子系统之间的竞争和协同，系统（宏观上）将**朝着**远离平衡的临界态发展。突变论研究了从系统一种稳定状态**跃迁**（渐变或飞跃）到另一种稳定状态的现象和规律，它认为一切形态的发生都归之于吸引子之间的**冲突**或斗争，通过控制吸引子的势可以改变系统的跃迁方式。超循环论解决了子系统之间如何利用物质、能量、和信息相互作用，并结合成为更紧密系统的**形式**问题。分形方法提供了系统的空间结构从简单走向复杂的演化或生成**过程**。混沌理论解释了对初始条件具有敏感依赖性的系统随时间动态非线性变化的**过程**。

我们提出的模型从内在驱动力和外在影响条件两个层面对群体智能的涌现进行了刻画，完整地阐述了群体智能涌现的条件、动力、途径、形式及过程。

### 6. 2. 面向特定问题的群体涌现模型

虽然提出和建立通用的群体智能涌现理论和模型具有重要的科学意义，但挑战性和难度也非常巨大，至少目前为止像我们的模型这样通用的模型还没有见到。因此，大多数群体智能模型或算法并不追求绝对的通用，它们大多是基于对特定的生命群体或社会活动进行观察，并汲取灵感而建立的[24]，甚至可以组合多种灵感源来应对高度复杂性问题[33]。

在[31]中，综述了众多群体智能模型，作者把模型分成四类：基于距离规则的模型（如 Boids[26]、PSO[9][18]）、由信息素相互作用形成的模型（如(ACO[11][6], BCA[16])、结合层次结构的模型([29][32])、以及源于实证研究的模型([27][34])。也有一些工作试图将不同类型的模型统一在一个框架中。例如，在[20]中，提出了一种智能优化系统的通用算法结构，通过设计模块化的框架，试图将所有算法（或技术）集成到一个统一的框架中，而不是建立一个新的模型。类似的， 在[15]中，提出了一种通用的群体智能的计算框架，它统一了几种典型的群体智能算法，如蚁群算法、粒子群算法等。

在这些模型中，有些偏重于个体的局部行为规则，个体的行为规则同时受到群体的制约或影响；有些则偏重于个体之间的交互模式或交互结构，认为群体智能的涌现主要源于个体

间的交互，个体在特定的交互模式或结构下调整自身的状态，参与合作或竞争，追求行为的最优化。尽管这些模型千差万别，但它们的底层逻辑都认为群体智能的涌现存在根本性的驱动力。例如，[10]的作者试图发现并解释群体智能出现的基本特征。在他们看来，群体智能是在环境约束下个体之间的相互作用中产生的，它是根据环境的动态变化而进化的，群体智能根植于环境的约束和动态。[5]的作者认为，涌现源于对逻辑或物理环境所施加的约束的微调或适应。[8]的作者认为，群体智能系统同样受到第一原理的支配，指出群体智能的第一原理是基于自然选择和进化律的，规模不变 Pareto 最优性规定了群体智能的演化方向。

在我们看来，以个体局部行为规则为核心的模型（我们可以称为内驱型或自驱型的）认为群体智能的主要驱动力来源于个体内部，即个体的内在利益或目标是驱动个体行为的根据，也是群体智能涌现的动力源泉；以交互模式或结构为核心的模型（我们可以称为外驱型或它驱型的）则认为群体智能的形成主要是受外部环境（包括环境中的信息）约束和驱动的，外部约束一方面影响了个体的行为，另一方面也规定了群体智能的形态。

但无论是内驱型的还是外驱型的模型都没有找准驱动群体智能涌现的根源，因为在不同的模型里，内驱因素和外驱因素并不相同，这些因素往往是先验的，是根据所期望涌现的群体智能特征而预先定制的。而在我们的模型里，群体智能是个体内部矛盾分布的反映，群体智能的产生和进化根源于个体内部矛盾的发展，群体智能的形态受到个体在环境中的交互的影响和制约。我们的模型不仅将内驱因素和外驱因素融合在了一起，并且将内驱因素和外驱因素归结为所有群体中都存在的因素，即内部矛盾与外部竞争，这使得我们的模型具有了普遍性和一般性。

## 7. 总结和讨论

人们在群体智能领域已进行了数十年的研究和探索，期望能找到涌现的内在原因并建立通用的涌现模型。本文提出了一种以矛盾为中心的群体智能涌现模型。在该模型中，个体的内在矛盾是个体存在和发展的根本，是驱动个体行为的内在动力，是决定个体性质的根本原因，同时，个体在群体中通过环境联系在一起，个体之间因为竞争环境资源而交互，并因此影响个体内在矛盾的演化。个体的联系及交互会形成具有整体性的组织结构或行为模式，这表明群体涌现出了群体智能。群体智能本质上可以看成是个体矛盾在群体中的分布状况的综合反映。

我们的模型既能刻画和模拟具有整体性的组织结构的涌现，又能刻画和模拟具有全局性的行为模式的涌现。从实验中可以看出，本文的模型还非常简单，在展示群体智能的涌现时基本上都不需要通过复杂的计算就可以实现。

尽管如此，本文提出的模型依然还存在一些限制，这也是我们将来要进一步探讨和完善的地方。例如，矛盾都是预先确定的。尽管个体在交互时会产生新的矛盾，但为了简单起见，我们不得不假设新产生的矛盾是可预见的，这样才能预先定义矛盾内部的博弈矩阵。其次，从概念上讲，有了内部矛盾，就应该有外部矛盾。个体与环境之间以及个体与个体之间也会存在矛盾，即外部矛盾。但个体间的外部矛盾从概念上讲很难看成是矛盾，因为它们更关注的是个体之间为了生存而产生的斗争，而斗争本质上都是因为竞争和占用环境资源而引起的交互产生的，因此我们在模型中没有刻意强调外部矛盾，而是把交互作为形成群体的更基础的概念来对待。这也是因为个体与外界环境和其它个体之间的外在联系是动态的，外部矛盾不像内部矛盾那么稳定，而个体为了生存所需占用的环境资源是相对稳定的，因此，在模型中强调交互而不是外部矛盾可以更好地刻画个体和群体。第三，现有的模型仅考虑了个体层面的矛盾，其实，在群体层也可能存在矛盾，例如社会矛盾，虽然说这些矛盾本质上也是从个体层的矛盾中涌现出来的。

在下一阶段，我们将对模型进行扩展，支持前面提及的各种动态性质，例如，允许个体产生新的矛盾，允许群体中生成新的个体，甚至允许群体中存在不同层次的矛盾。另外，当前模型中对个体行为和交互的描述还很初步，在将来，我们将为个体及其交互建立一种数学演算系统，从而可以严谨地推到群体智能的涌现。

## 参考文献